# An Approach for Automatic Construction of an Algorithmic Knowledge Graph from Textual Resources

Jyotima Patel[1,2] and Biswanath Dutta[3]

[1] DRTC, Indian Statistical Institute, Bangalore, India, jyotima@drtc.isibang.ac.in
[2] Department of Library and Information Science, Calcutta University, Kolkata, India
[3] DRTC, Indian Statistical Institute, Bangalore, India, bisu@isibang.ac.in

**Abstract**
There is enormous growth in various fields of research. This development is accompanied by new problems. To solve these problems efficiently and in an optimized manner, algorithms are created and described by researchers in the scientific literature. Scientific algorithms are vital for understanding and reusing existing work in numerous domains. However, algorithms are generally challenging to find. Also, the comparison among similar algorithms is difficult because of the disconnected documentation. Information about algorithms is mostly present in websites, code comments, and so on. There is an absence of structured metadata to portray algorithms. As a result, sometimes redundant or similar algorithms are published, and the researchers build them from scratch instead of reusing or expanding upon the already existing algorithm. In this paper, we introduce an approach for automatically developing a knowledge graph (KG) for algorithmic problems from unstructured data. Because it captures information more clearly and extensively, an algorithm KG will give additional context and explainability to the algorithm metadata.

## 1 Introduction

Scientific knowledge is distributed and maintained through academic literature published in, for example, conference proceedings, journal articles, and workshop proceedings. It is a challenging task for researchers to keep track of innovations like proposals for new frameworks, algorithms, and software with such an increased number of publications. Algorithms (where an algorithm is a step-by-step strategy to tackle any issue) are published in areas ranging from Mathematics to Geo-sciences and Computer Science [1]. Algorithms in the scientific literature are expressed as flowcharts, pseudo-codes, or computer programs. As algorithms are written in a step-by-step manner, understanding a scientific problem becomes simple. Studying an algorithm also helps to know the processes and techniques used to solve a scientific problem. In the academic community, algorithms are often used to communicate the goal, technique, and procedures taken towards solving a problem, whereas practical people more often look for the implementation of the algorithm [2]. There is no system available where the researchers can find all the major algorithms, though there are initiatives like The Stony Brook

Algorithm Repository[1] [2], Algowiki[2] [3], and Wikidata[3] . The main limitations of these repositories are that they provide a minimal description of the algorithms, they also lack a proper search facility and browsing for algorithms in these repositories is a tedious and time-consuming process (the repositories are further elaborated on in section 2). Because of these issues, many important algorithmic works often go unnoticed. Also, as stated above, since the algorithms are described minimally, searching for them is a challenge. The algorithm metadata in most cases is either missing or provided in a very minimal way. The extraction of the metadata is a challenging task because the information is usually located within the scholarly or other textual resources, such as the registries and repositories, like Algowiki and Stony Brook Algorithm Repository. Gathering the metadata manually from these resources is a tedious and time-consuming process. One of the main focuses of the current study is to provide an approach for extracting
the algorithmic metadata from textual resources. As a first step, we consider the Stony Brook Algorithm Repository as a potential source for extracting the metadata. Further, we transform these metadata automatically into a knowledge graph (KG) (*a manifestation of an intelligent web of Data informed by an ontology* [4] [5]). The transformation of algorithm metadata into a KG helps in representing the algorithms and their relations with high dependability, logic, and reusability. Apart from providing better search and retrieval, the KG identifies related information and helps in the comparison of algorithms. The KG can also be used as a referential model for the automatic extraction of information from the scientific literature.
The primary contributions of this work are:

- a methodology for the automatic creation of a knowledge graph for algorithmic problems.
- an approach for the automatic extraction of algorithmic problems and associated data from textual resources.
- generation of a knowledge graph using the extracted data.

The rest of the paper is organized as follows: Section 2 discusses the existing algorithm repositories and provides a comparison between them; section 3 gives an overview of the KG development approach and the steps involved in it; section 4 discusses the data extraction i.e the steps involved in the extraction of the data and the data processing; section 5 discusses the KG creation process from the processed data and also the tools used; section 6 provides some SPARQL queries to validate the KG; section 7 discusses the relevant related works and Section 8 concludes the paper and provides future research directions.

## 2  Existing Algorithm Repositories

Algorithm repositories are storage areas for algorithms, where algorithms are often stored with minimal metadata. We found three dedicated algorithm repositories on the web. The Stony Brook Algorithm Repository provides an exhaustive collection of 75 fundamental algorithmic problems along with their implementations. The algorithms in this repository are categorized into two groups: language and by problem [2]. Each algorithm has its own separate web page containing information like input, output, input description, implementations, related problems, and so on.

---

[1] https://algorist.com/algorist.html
[2] https://wiki.algo.is/
[3] https://www.wikidata.org/

Algowiki is an online encyclopedia of algorithms. It contains a list of algorithms on a wiki page arranged in alphabetical order. Algowiki consists of algorithms from the field of mathematics with some features and properties. It is a wiki dedicated to competitive programming [3]. The algorithms in the Stony Brook algorithm repository are held in one web Server, whereas in Algowiki once you click on the desired algorithm, the site is redirected to either a Wikipedia page or a website that holds the description of the algorithm with minimal metadata. In Algowiki, only the URL of each algorithm is stored [3].

Wikidata is a knowledge base that acts as central storage for structured data of its Wikimedia sisters [6]. Initiatives are taken by Wikidata to model algorithms where they use generic properties to describe algorithms and then interlink the various other algorithms as sub-classes or instances. The Wikidata repository consists of items, each having a label and description [6]. When compared with the Stony Brook repository, Wikidata provides a more generalized description. For instance, the Convex Hull problem is described with metadata elements, such as title, description, sub-class, and main category. The metadata like input, output, and related problems are not provided. Wikidata has modeled the algorithms in a broad manner. The metadata elements present in Wikidata are in a structured format. Wikidata provides query service through the SPARQL endpoint[4].

Figure 1: Algorithmic description in Stony Brook Algorithm Repository.

Taking note of the data extraction for the KG creation, Wikidata provides a seamless service to extract data in many formats like csv, json, xml, and many more. Wikidata provides structured data by just executing SPARQL queries that do not fit the scope of our current work. Whereas, the

---
[4] https://query.wikidata.org/

Stony Brook Algorithm repository challenges us to extract the unstructured data from its website and process it to the set format. Algowiki does not meet our requirements as it describes the algorithms minimally. It is more like a registry and not a repository as the information is not held in the Algowiki server but spread across various third party websites. Whereas in Stony Brook Repository apart from algorithmic problems various associated entities like implementations and related problems are also described. Hence, in this work, our emphasis lies on the Stony Brook Algorithm Repository for data extraction. Figure 1 shows a representational algorithmic description from the repository.

## 3  Knowledge Graph Development Approach

In this work, we build a KG for describing algorithmic problems and associated entities, such as software implementations, recommended resources, persons, and so forth. We present here a general approach toward the KG creation for algorithms and their relations. Figure 2 shows the important steps that we follow.

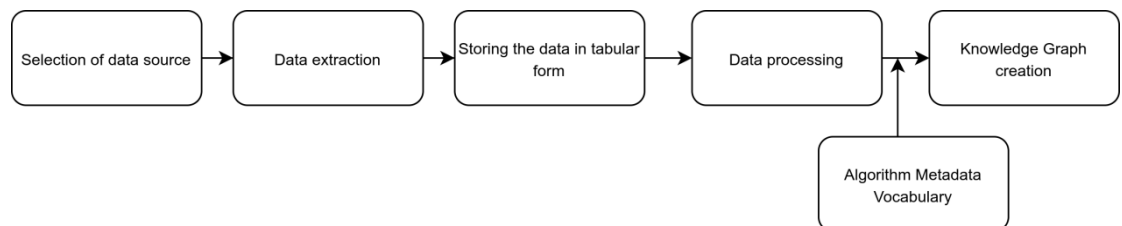

Figure 2: Steps for KG Creation.

**Step 0- Selection of data source.** The initial step is to identify the objective data source for data extraction, In section 2 comparisons are drawn among the accessible Algorithm Repositories. The Stony Brook Algorithm Repository is recognized as the potential source to extract the data for generating the KG.

**Step 1- Data extraction.** The Stony Brook Algorithm Repository has 75 separate web pages for its algorithms containing information like input, output, problem statement, description, related problems, implementation, rating, and so on. There are 10 metadata elements about each algorithm that interest us for extraction. Using an HTML parser and pattern matching the metadata present in each webpage is extracted and stored in a convenient data structure. In this work Python3, Dictionary and List are used as our preferred data structure.

**Step 2- Storing the data in tabular form.** In the above step, the data is stored in a Python variable but it needs to be exported to use it outside the Python environment. A Python package called Pandas is used to export it in tabular form.

**Step 3- Data processing.** The exported data file needs to be further processed in such a way that KG transformation can be done easily. In the present data file, there are metadata elements that have one to many relationships and are represented in a single column. These need to be separated into multiple columns. Further, there are metadata elements that are combined, for instance for an algorithm problem recommended books information column contains the book title along with the author's information as a single entity, which needs to be segregated into multiple columns.

**Step 4- Knowledge Graph creation.** The above data is taken as input and utilizing the Algorithm Metadata Vocabulary (AMV) [7] and MappingMasterDSL [8] the data is transformed into a KG. The details in regard to the KG development are discussed in section 5.

## 4 Data Extraction and Processing

As discussed in section 2, for the current study, we selected Stony Brooks Algorithm Repository as our target repository to extract data. The Repository categorizes the algorithmic problems into two main classes: problems and language. On its home page ("https://algorist.com/algorist.html"), all the algorithmic problems are listed. Each problem in the list points to an HTML page where a detailed depiction of that algorithmic problem is present. Figure 3 shows a list of algorithmic problems. Information about each algorithmic problem, e.g., Convex Hull, String Matching, and Text Compression is to be extracted. The algorithmic problems are further grouped under seven broad problem types, for example Combinatorial Problems, Graph Problems, and Computational Geometry. Figure 3 shows a few of them.

The process of extraction of information from the website is completely automated. To automate the extraction and process the information in the required format the following Python libraries were used: BeautifulSoup, Selenium, Urllib, Regex, and Pandas. Among these BeautifulSoup along with Urllib is used to extract the relevant text from HTML pages. Webdriver from the Selenium library is used for navigating to different HTML pages. Lastly, Pandas and Regex are used to tabulate and clean the extracted data.

### 4.1 Data Extraction

In this section, the steps for extracting information from the Stony Brook Algorithm repository are discussed. The pseudo-code for data extraction is given as Algorithm 1 (the source code is available on GitHub[5] ).

Due to the space problem, only two examples are provided. The extracted values as shown in example 1 are appended to their corresponding keys in the dictionary created in step 3.

**Step 0- Importing the Python libraries.** Started with importing the Python libraries, regex, urllib, bs4, selenium and Pandas.

**Step 1- Links stored in a list.** The selenium web driver is used to open the URL "https://algorist.com/algorist.html" and all the links present on the homepage (as shown above in Figure 3 each algorithmic problem is a link to an HTML page) are stored in a list (a Python data structure).

Figure 3: Homepage of the Stony Brook Algorithm Repository

**Step 2- Filtering the URLs of algorithmic problems.** In the above step all the URLs present on the homepage are stored, but our interest lies in the URLs of each algorithmic problem. Hence, an empty list 'AlgorithmicProblem' is created and all the URLs starting with 'https://algorist.com/problems/' are stored since the URL for all algorithmic problems starts with the mentioned pattern. Regex is used to filter out the desired URLs.

---

[5] https://github.com/biswanathdutta/amv

**Step 3- Dictionary created for storing metadata.** By visual inspection of the website it is clear that each algorithmic problem has ten metadata elements. Hence, a Python dictionary is created with keys: problem, problem type, input image, output image, input decription, problem statement, description, implementations, recommended books and related problems. Initially all the keys are assigned an empty list.

**Step 4- Data population.** This step uses links that were stored in step 1 to populate the value in the dictionary corresponding to key problem type. Regex is used to filter out the links that begin with 'https://www.algorist.com/sections' and store it in a list. Since, all the broad categories, e.g., Data Structures, Numerical Problems, and Combinatorial Problems, have URLs starting with 'https://www.algorist.com/sections'.

**Step 5- Metadata extraction.** In this step, the metadata of each algorithmic problem is extracted. BeautifulSoup 'soup'(it contains HTML code in a hierarchical manner which is easy to access) object is created which contains the HTML script of each algorithmic problem. This is achieved by iterating over the list 'AlgorithmicProblem'. For instance, the title of the algorithmic problem is available in the 'h1' tag of the HTML page. Using the soup.select function it is accessed, there are multiple implementations present and each has its 'name', 'url', and 'rating'. This information is extracted as shown in example 1.

**Example 1.** *name1 | url1 | rating1 | implementation1 language 1 \n name2 | url2 | rating.*

**Step 6- Exporting the output as csv.** A Pandas dataframe is created from the main dictionary that holds all the information and this data frame is exported as a csv file.

---

**Algorithm 1** Extracting data

**Input**: URL of the website
**Output**: algo_dict

```
 1: Import libraries: sys, regex, urllib, BeautifulSoup, selenium, pandas, numpy.
 2: driver = Instantiated selenium webdriver and open url "https://algorist.com/algorist.html"
 3: store algorithms = list of URLs of algorithmic problems using driver
 4: store sections = list of URLs of Types of algorithmic problems
 5: driver close
 6: create algo_dict = Dictionary with keys problem, problem_url, problem_type, input_image, output_image, input_decription,
    problem_statement, description, recommended_books, related_problems and values = empty list.
 7: for url in sections do
 8:     driver = Instantiate selenium webdriver and open url
 9:     store c = Number of algorithmic problems in this type
10:     algo_dict["problem_type"].extend(c times type name)
11:     driver close
12: end for
13: for url in algorithms do
14:     soup = Instantiate BeautifulSoup soup object and open url
15:     for key in algo_dict do
16:         algo_dict[key].append(relevant metadata from soup)
17:     end for
18: end for
19: driver = Instatntiate selenium webdriver and open "https://algorist.com/algorist.html"
20: store languages = list of URLs with languages of the algorithmic problems using driver
21: driver close
22: create dict_75 = dictionary with keys as the elements in algo_dict["problem_url"]
23: for url in languages do
24:     store language_name = name of the current language
25:     driver = Instatntiate selenium webdriver and open url
26:     Add implementation, rating, implementation_url, and name of language seperated by " I " to corresponding key whenever
        a problem_url is present in this page. Add a "\n" at the end of the string
27:     driver close
28: end for
29: store algos_dict['implementations'] = values in dict_75
30: Export algos_dict as a csv file using pandas
```

Algorithm for information extraction

## 4.2 Data Processing

The exported data in csv (as discussed above), needs to be processed and converted into a form that fits with the Algorithm Metadata Vocabulary (AMV) model (discussed in section 5).The data will be utilized to create a KG and in a KG each entity is represented in a uniquenode [9]. In the KG, a node can be any object, place, or person and the edge defines the relationship between nodes [10]. In the exported data, some data points are merged together and do not represent a unique entity. Hence, the processing is required for some columns e.g. implementation and related problems.

The columns like implementation, related problems, and recommended books have more than one element. Figure 4 shows the columns implementation, recommended books, related problems, and so on. As visible in the Figure 4 there is information that needs to be in different columns. For instance, the implementation column has multiple entities. Firstly, each individual implementation needs to be separated. Each individual implementation also has its name, rating, and url, this information also needs to be split. Similar processing is required for the columns recommended_books, related_problems, and implementation_in_languages.

| implementations | recommended_books | related_problems |
|---|---|---|
| libc++ \| (rating 10) \| https://llvm.org/svn/ll... | Data Structures and Algorithm Analysis in C++ ... | Priority Queues \| https://algorist.com/problem... |
| libstdc++ \| (rating 10) \| https://gcc.gnu.org/... | Data Structures and Algorithm Analysis in C++ ... | Dictionaries \| https://algorist.com/problems/D... |

Figure 4: Shows the raw data after extraction

The details of processing the data are as follows:

**Step 0- Importing the libraries and loading the data.** Started by importing the Python libraries, numpy, Pandas and regex and the csv file is loaded which contains the extracted data.

**Step 1- Processing the implementations.** In this step, the focus is on the implementation column, the multiple implementations were combined using '\n'. The same is used to split each implementation to get the maximum number of implementations present and create that many columns with suffix (eg. Implementation 1, Implementation 2). Further, each implementation is appended in separate columns that were created. Each implementation has its title, rating, and url which are combined using '|'. The same is used to split them and add them to separate columns.

**Step 2- Processing the related problems.** In this step, the focus is on the related problem column. There are multiple related problems combined with '/n' and each problem has its name and url combined using '|'. A similar approach to the previous step is followed to process
this column.

**Step 3- Processing the recommended books.** In this step, the focus is on the recommended book column, the book name contains the title of the book along with author names. The book title and author's name are separated by 'by' string. For books with multiple authors, each author is separated by 'and' and ','. Following the similar approach as above the books are split into multiple columns and the related information like author name, book url are split from each book.

**Step 4- Export the processed data.** The final data frame is exported into a .xlsx file.

Figure 5 shows the recommended books column before and after processing. As visible in Figure 5 after extraction, each recommended book column has its title, url, and authors. After processing, each recommended book column is split and their authors, title and URLs are also present in different columns.

| recommended_books |
|---|
| Practical Algorithms for Programmers by A. Binstock and J. Rex \| https://www.amazon.com/exec/obidos/ASIN/020163208X/thealgorith01-20?tag=algorist-20 Handbook of Algorithms and Data Structures by G. Gonnet and R. Baeza-Yates \| https://www.amazon.com/exec/obidos/ASIN/0201416077/thealgorith01-20?tag=algorist-20 The Design and Analysis of Computer Algorithms by A. Aho and J. Hopcroft and J. Ullman \| https://www.amazon.com/exec/obidos/ASIN/0201000296/thealgorith01-20?tag=algorist-20 Regular Algebra and Finite Machines by J. H. Conway \| https://www.amazon.com/exec/obidos/ASIN/0412106205/thealgorith01-20?tag=algorist-20 |

| recommended_book_1_name | recommended_book_1_author_1 | recommended_book_1_author_2 | recommended_book_1_author_3 | recommended_book_1_au... |
|---|---|---|---|---|
| Practical Algorithms for Programmers | A. Binstock | J. Rex | nan | |

Figure 5: Shows the recommended books information before and after processing.

## 5 Knowledge Graph Creation

The processed data was received in a .xlsx file with 75 rows and 163 columns. This data is taken as input for the KG development. In the current study, Algorithm Metadata Vocabulary (AMV[6]) is used as a schema for the KG. MappingMaster ($M2$ ) is used to transform the data
into a KG. AMV is a metadata vocabulary for describing algorithms, algorithm problems, and related entities, like software code. The vocabulary is available as an OWL ontology. It can be directly used by anyone interested to create and publish algorithm metadata as a KG, or to provide metadata service through the SPARQL endpoint [11]. We have used it as a schema for the production of the algorithmic KG. MappingMaster (an open-source Java library to transform the content of spreadsheets to OWL axioms) has GUI support and is available as a plugin called Cellfie for Protege Desktop [12]. MappingMaster has domain-specific language (DSL) [8]. For the current work, we used the Cellfie plugin and also the DSL language for developing mapping rules for transforming the metadata available in a spreadsheet (as mentioned above) into KG. A snippet of the developed mapping rules expressed in ($M^2$) DSL language is shown in Table 1.

---

[6] https://w3id.org/amv

| | A | B | C | D | E | F | G | H |
|---|---|---|---|---|---|---|---|---|
| 1 | problem | problem_url | problem_type | input_image | output_image | input_description | problem_statement | description |
| 2 | Dictionaries | https://algoris | Data_Structures | https://algoris | https://algorist. | A set of \(n\) recor | Build and maintain a | Excerpt from Th An essential pie |
| 3 | Priority Queues | https://algoris | Data_Structures | https://algoris | https://algorist. | A set of records w | Build and maintain a | Excerpt from Th If your applicatic However, if you |
| 4 | Suffix Trees and | https://algoris | Data_Structures | https://algoris | https://algorist. | A reference string | Build a data structur | In its simplest in A trie is a tree st Tries are useful Starting with the A suffix tree is s |
| 5 | Graph Data Stru | https://algoris | Data_Structures | https://algoris | https://algorist. | A graph \(G\). | Give an efficient, fle | Excerpt from Th Building a good |
| 6 | Set Data Structu | https://algoris | Data_Structures | https://algoris | https://algorist. | A universe of obje | Represent each sub | Excerpt from Th When each subs Your primary alt |

Figure 6: A glimpse of the dataset

Table 1 provides a snippet of the mapping rule for the algorithm problems and their corresponding types, identifier, representational depiction of input and output, input description for the problem, problem description, and implementation details.

```
Individual: @A*(rdfs:label=(@A*))
Types: @C**
Facts: dcterms:title @A*,
       dcterms:identifier @B*,
       inputImage @D*,
       outputImage @F*,
       inputDescription @H*,
       problemDescription @I*,
       excerpt @J*,
       hasImplementation @K*
```

Table 1: Snippet of the mapping rules represented in *M*2 DSL language

The produced KG consists of 1494 individuals and 9706 axioms in addition to 59 classes, 44 object properties, and 50 data properties coming from the AMV ontology. The KG is available as an RDF dump and can be downloaded from GitHub[7].

## 6 SPARQL Queries

To validate the produced KG, we have conducted several SPARQL queries. The queries are centered around the algorithm problems and their e.g. relations, implementation, and related information like implementation language, platform, loop types, and data structure. Some of the queries are: retrieve all the problems related to the sorting problem, with their corresponding type and their implementation details like problem statement, implementation URI and implementation language (Q1), and retrieve the implementations of Eulerian Cycle problem in C++ programming language (Q2) and retrieve the related algorithms for text compression problem along with their looping structure (Q3). Table 2 provides the SPARQL representation for the first query(Q1). Figure 7 displays the query result in a graph produced using a browser-based application Gruff (https://allegrograph.com/products/gruff/). The successful execution and the retrieval of desired results for the various queries centred on algorithm problems prove the efficacy of KG.

---
[7] https://github.com/biswanathdutta/amv

```
PREFIX dct: <http://purl.org/dc/terms/>
PREFIX amv: <https://w3id.org/amv#>
PREFIX rdfs:<http://www.w3.org/2000/01/rdf-schema#>
SELECT DISTINCT ?problem ?type ?prob desc ?y
?impl uri ?impl language
WHERE amv:Sorting dct:relation ?problem.
?problem a ?type ; amv:problemDescription
?prob desc ; amv:hasImplementation ?y .
?y amv:inProgrammingLanguage
?impl language; dct:identifier ?impl uri.
```

Table 2: Shows SPARQL query representation for Q1

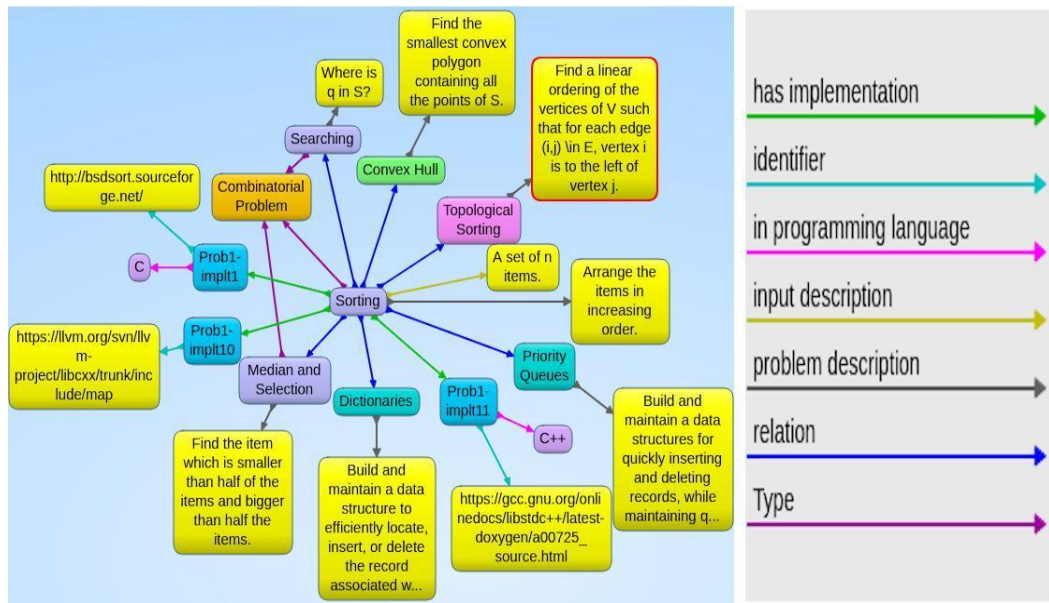

Figure 7: Showing the result in a graph for the query in Table 2

## 7 Related Works

Addressing all encompassing and factual knowledge utilizing RDF and Linked Data is quite feasible [13]. There are industrial innovations like Thomson Reuters KG Feed for the Financial Services market [13] and BBC's KG for their operations with content [9]. Google's KG to make web searches more intelligent and augment the results with information relevant to the query [14]. In academia, much work is focused on addressing the bibliographic metadata, while machine readable portrayal of scientific information in academic writing has not received much consideration [13]. There are few methodologies that focus on scientific literature. The Artificial Intelligence Knowledge Graph(AI-KG) is a large-scale automatically generated KG that depicts research entities. It uses deep learning techniques to extricate elements and relations from scientific text [15].

The Open Research Knowledge Graph (ORKG) [13] contributes toward representing scholarly knowledge semantically with KGs. ORKG not only contains the bibliographic metadata like authors, references, but also contains semantic depiction of scholarly literature like problem statement, approach and implementation. Both AI-KG and ORKG utilize deep learning techniques for extraction from the academic literature but KG of scientific software metadata focuses on external code repositories, readme files and documentation of software [16]. It focuses on metadata categories like description, installation instructions, execution, and citation for extraction. Machine Learning techniques were employed to gather the data whereas in our work we used pattern matching to gather data. For KG development a list of programme items is scrapped from a target software registry (e.g., Zenodo). Then, for each item, its version data is obtained, extract all code repository links, and download the complete text of its readme file. SOMEF parses the readme file, and the findings are integrated and aggregated into a knowledge Graph [1].

OKG-Soft is also one effort towards creating a KG for scientific software metadata in a machine readable manner. OKG-Soft includes an ontology designed to describe software and the specific data formats it uses and publish software metadata as an open KG, linked to other Web of Data objects [17]. The capture of metadata is based on their previous work OntoSoft [18].

The one ongoing effort towards KG is OpenAIRE, which considers many scientific arti-facts like research literature, research software, and research data. It also includes metadata records about organizations involved in the research life-cycle, such as universities, research organizations, and funders [19]. Graph4code is another work that focuses on the program code, the metadata in this is extracted by the code documentation, forum discussions and then mapped into a knowledge graph, this work employs extensive use of named graphs in RDF to make the knowledge graph extensible [20]. The work focuses on and captures the semantics of Python codes whereas our work focuses on algorithms that can be implemented in different programming languages. Despite covering such diverse metadata records from various academic literature and external sources, the idea of depicting the metadata of algorithms and related entities and their relations is not considered previously. The current work may be considered a pioneer in this regard.

## 8  Conclusion

Algorithms published in the scientific literature are crucial to comprehend and reuse to better understand the information. There is a surge in the number of research publications made available each year. This creates a necessity to make algorithm searches more effective and personalized. Algorithms should be treated as independent digital objects like research data, research articles and in recent times software and ontologies [21]. In this work, we have presented a novel approach for automatically creating a KG by extracting the data of 75 different algorithmic problems along with their relations and a methodology from the textual resource like Stony Brook algorithm repository. The work explores new metadata categories related to algorithmic problems to better comprehend problems and reuse. Such categories are related problems (a similar algorithmic problem) and recommended books (to better understand the problem and get detailed explanations). In continuation to the current work, we aim to make a comparative analysis of the designed KG approach with state of the art related approaches as used in the creation of knowledge graphs, such as AI-KG and ORKG. The present KG was developed primarily based on a single resource i.e., the Stony Brook algorithm repository. In the future, we aim to extend the KG by extracting the information from several other sources, such as scientific literature, repositories (e.g., GitHub, NIST Dictionary of Algorithms and Data Structures [22]), and online discussion forums. For this purpose, we aim to focus on developing a more generic and robust framework of information extraction.

**References**

[1] A. Kelley and D. Garijo, "A framework for creating knowledge graphs of scientific software metadata," *Quantitative Science Studies*, p. 1–37, Nov 2021.

[2] S. S. Skiena, "Who is interested in algorithms and why? lessons from the stony brook algorithms repository," *In: Proc. of WAE'98*, pp. 204–212, 1998.

[3] V. Voevodin, A. Antonov, and J. Dongarra, "Why is it hard to describe properties of algorithms?" *Procedia Computer Science*, vol. 101, p. 4–7, Dec 2016.

[4] K. U. Idehen, "Linked data, ontologies, and knowledge graphs," 2020. [Online].Available:https://www.linkedin.com/pulse/linked-data-ontologies-knowledge -graphs-kingsley-uyi-idehen/

[5] M. DeBellis and B. Dutta, "The covid-19 codo development process: an agile approach to knowledge graph development," 2021.

[6] "Wikidata." [Online]. Available: https://www.wikidata.org/wiki/Wikidata:Main Page

[7] B. Dutta and J. Patel, "AMV: Algorithm Metadata Vocabulary," *arXiv:2106.03567 [cs]*, Jun 2021, arXiv: 2106.03567. [Online]. Available: http://arxiv.org/abs/2106.03567

[8] M. O'Connor, C. Halaschek-Wiener, and M. Musen, "Mapping master: A flexible approach for mapping spreadsheets to owl," *9th International Semantic Web Conference (ISWC), Shanghai, China*, vol. 6497, p. 208, Nov 2010, journalAbbreviation: 9th International Semantic Web Conference (ISWC), Shanghai, China.

[9] T. Petkova, "The knowledge graph and the enterprise," Aug 2018. [Online]. Available: https://www.ontotext.com/blog/the-knowledge-graph-and-the-enterprise/

[10] I. C. Education, "What is a knowledge graph," November 2021. [Online]. Available: https://www.ibm.com/cloud/learn/knowledge-graph

[11] B. Dutta and M. Debellis, "CODO: An ontology for collection and analysis of covid-19 data," *In Proc. of 12th Int. Conf. on Knowledge Engineering and Ontology Development (KEOD)*, pp.76–85, 2-4 November 2020.

[12] M. Musen, "The prot´eg´e project," *AI Matters*, vol. 1, p. 4–12, Jun 2015.

[13] S. Auer, M. Stocker, L. Vogt, G. Fraumann, and A. Garatzogianni, "Orkg: Facilitating the transfer of research results with the open research knowledge graph," Research Ideas and Outcomes, vol. 7, 2021.

[14] C. Harding, "Semantic data platforms come of age — linkedin." [Online]. Available: https://www.linkedin.com/pulse/semantic-data-platforms-come-age-chris-harding/

[15] J. Tsay, A. Braz, M. Hirzel, A. Shinnar, and T. Mummert, "Aimmx: Artificial intelligence model metadata extractor," in *Proceedings of the 17th International Conference on Mining Software Repositories*. ACM, Jun 2020, p. 81–92. [Online]. Available: https://dl.acm.org/doi/10.1145/3379597.3387448

[16] A. Mao, D. Garijo, and S. Fakhraei, "Somef: A framework for capturing scientific software metadata from its documentation," p. 3037, Dec 2019.

[17] D. Garijo, M. Osorio, D. Khider, V. Ratnakar, and Y. Gil, "Okg-soft: An open knowledge graph with machine readable scientific software metadata," *15th International Conference on eScience (eScience)*, 2019.

[18] Y. Gil, V. Ratnakar, and D. Garijo, "Ontosoft: Capturing scientific software metadata," in *Proceedings of the 8th International Conference on Knowledge Capture*. ACM, Oct 2015, p. 1–4. [Online]. Available: https://dl.acm.org/doi/10.1145/2815833.2816955

[19] P. Manghi, C. Atzori, A. Bardi, J. Schirrwagen, H. Dimitropoulos, S. La Bruzzo, Y. Foufoulas, A. L¨ohden, A. B¨acker, A. Mannocci, M. Horst, M. Baglioni, A. Czerniak, K. Kiatropoulou, A. Kokogiannaki, M. Bonis, M. Artini, E. Ottonello, A. Lempesis, and F. Summann, "Openaire
research graph dump," Dec 2019.

[20] K. Srinivas, I. Abdelaziz, J. Dolby, and J. McCusker, "Graph4code: A machine interpretable knowledge graph for code," Feb 2020.

[21] B. Dutta, A. Toulet, V. Emonet, and C. Jonquet, "New generation metadata vocabulary for ontology description and publication," Metadata and Semantic Research, p. 173–185, 2017.

[22] P. E. Black, "Dictionary of algorithms and data structures," 1998. [Online]. Available: http://www.nist.gov/dads